\documentclass[sigconf, screen]{acmart}

\AtBeginDocument{%
  }

\usepackage{multirow}
\usepackage{pifont}
\usepackage{algorithm}
\usepackage{algpseudocode}
\usepackage{graphicx}
\usepackage{subcaption} 
\usepackage{libertinus}

\renewcommand\footnotetextcopyrightpermission[1]{}
\begin{document}

%%
%% The "title" command has an optional parameter,
%% allowing the author to define a "short title" to be used in page headers.
% \title{A Cross-Modal Multi-Region Invariant Features Learning Method for Single-Domain Generalized Object Detection}
\title{Boosting Single-Domain Generalized Object Detection via Vision-Language Knowledge Interaction}

% Transferring Fine-grained Vision-Language Knowledge for xxxx

% xxx: A Vision-Language 

% Boosting Single-domain Generalized Object Detection via Vision-Language Knowledge Interaction.

% Boosting 

% alignment Vision-Language alignment

% Learning invariant feature via xx for xx

% Feature-invariant learning method

% for Single-Domain Generalized Object Detection
%%
%% The "author" command and its associated commands are used to define
%% the authors and their affiliations.
%% Of note is the shared affiliation of the first two authors, and the
%% "authornote" and "authornotemark" commands
%% used to denote shared contribution to the research.

\author{Xiaoran Xu}
\affiliation{%
  \institution{School of Advanced Interdisciplinary Sciences, University of Chinese Academy of Sciences}
  % \city{Chaoyang Qu}
  \state{Beijing}
  \country{China}
}
\email{xuxiaoran22@mails.ucas.ac.cn}

\author{Jiangang Yang}
\authornote{Corresponding authors.}
\affiliation{%
  \institution{Institute of Microelectronics of the Chinese Academy of Sciences}
  % \city{Chaoyang Qu}
  \state{Beijing}
  \country{China}
}
\email{yangjiangang@ime.ac.cn}

\author{Wenyue Chong}
\affiliation{%
  \institution{School of Advanced Interdisciplinary Sciences, University of Chinese Academy of Sciences}
  % \city{Chaoyang Qu}
  \state{Beijing}
  \country{China}}
\email{chongwenyue22@mails.ucas.ac.cn}

\author{Wenhui Shi}
\affiliation{%
  \institution{Institute of Microelectronics of the Chinese Academy of Sciences}
  % \city{Chaoyang Qu}
  \state{Beijing}
  \country{China}
}
\email{shiwenhui@ime.ac.cn}

\author{Shichu Sun}
\affiliation{%
  \institution{School of Advanced Interdisciplinary Sciences, University of Chinese Academy of Sciences}
  % \city{Chaoyang Qu}
  \state{Beijing}
  \country{China}
}
\email{sunshichu22@mails.ucas.ac.cn}

\author{Jing Xing}
\affiliation{%
  \institution{Institute of Microelectronics of the Chinese Academy of Sciences}
  % \city{Chaoyang Qu}
  \state{Beijing}
  \country{China}
}
\email{xingjing@ime.ac.cn}

\author{Jian Liu}
\authornotemark[1]
\affiliation{%
  \institution{Institute of Microelectronics of the Chinese Academy of Sciences}
  % \city{Chaoyang Qu}
  \state{Beijing}
  \country{China}
}
\email{liujian@ime.ac.cn}

%%
%% By default, the full list of authors will be used in the page
%% headers. Often, this list is too long, and will overlap
%% other information printed in the page headers. This command allows
%% the author to define a more concise list
%% of authors' names for this purpose.
\renewcommand{\shortauthors}{Xu et al.}

%%
%% The abstract is a short summary of the work to be presented in the
%% article.

\begin{abstract}
Single-Domain Generalized Object Detection~(S-DGOD) aims to train an object detector on a single source domain while generalizing well to diverse unseen target domains, making it suitable for multimedia applications that involve various domain shifts, such as intelligent video surveillance and VR/AR technologies. With the success of large-scale Vision-Language Models, recent S-DGOD approaches exploit pre-trained vision-language knowledge to guide invariant feature learning across visual domains. However, the utilized knowledge remains at a coarse-grained level~(e.g., the textual description of adverse weather paired with the image) and serves as an implicit regularization for guidance, struggling to learn accurate region- and object-level features in varying domains. In this work, we propose a new cross-modal feature learning method, which can capture generalized and discriminative regional features for S-DGOD tasks. The core of our method is the mechanism of Cross-modal and Region-aware Feature Interaction, which simultaneously learns both inter-modal and intra-modal regional invariance through dynamic interactions between fine-grained textual and visual features. Moreover, we design a simple but effective strategy called Cross-domain Proposal Refining and Mixing, which aligns the position of region proposals across multiple domains and diversifies them, enhancing the localization ability of detectors in unseen scenarios. Our method achieves new state-of-the-art results on S-DGOD benchmark datasets, with improvements of +8.8\%~mPC on Cityscapes-C and +7.9\%~mPC on DWD over baselines, demonstrating its efficacy.

\end{abstract}

\maketitle
\begin{figure}[!t]
    \centering
    \includegraphics[width=\linewidth]{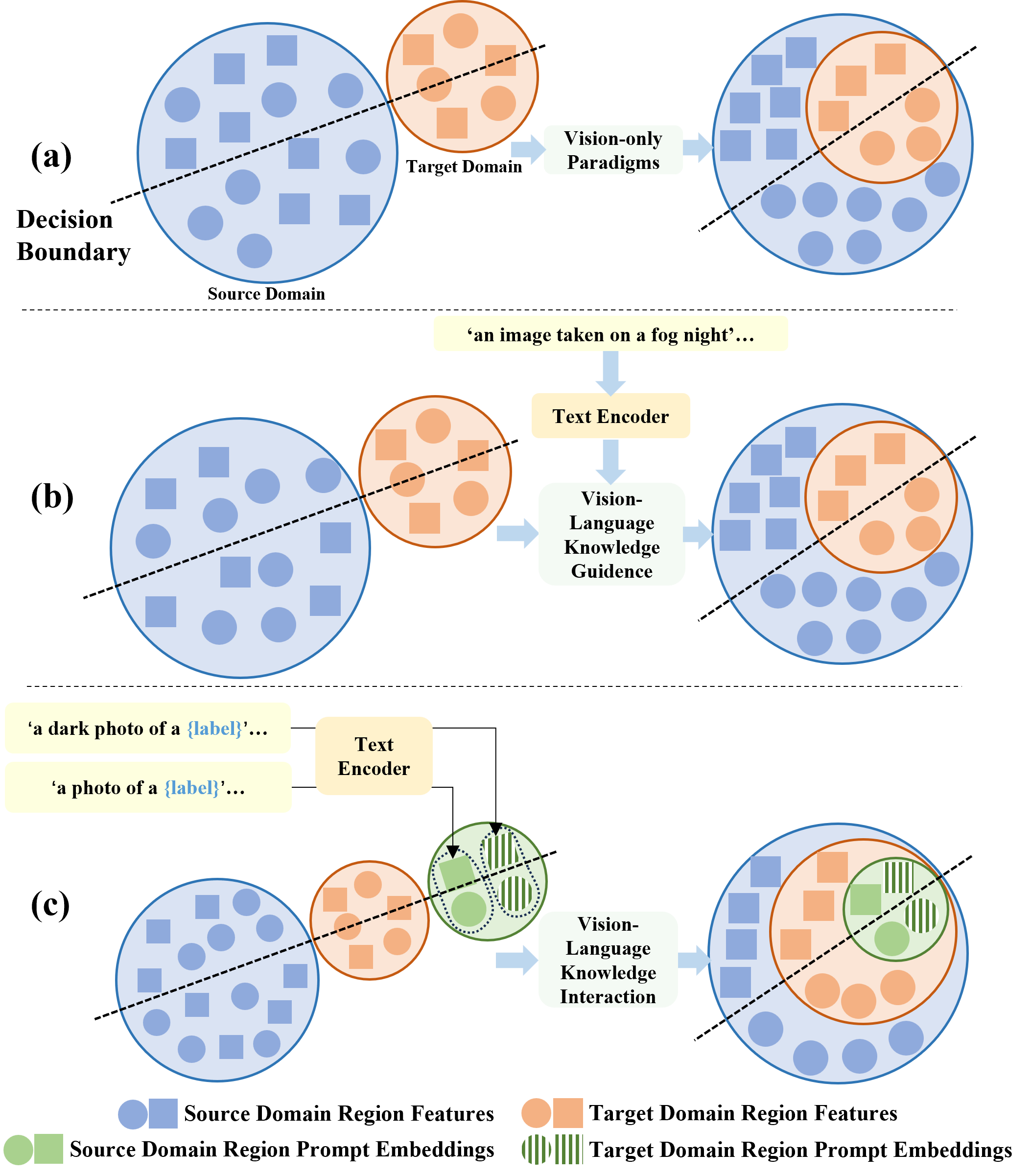}
    \caption{This figure provides a high-level comparison of three distinct strategies in S-DGOD.  Different shapes in the figure mean different regions. (a) Vision-only methods; (b) VLM-based methods; (c) Our method. 
}
    \label{Figure 1}
\end{figure}

\section{INTRODUCTION}
% \textcolor{red}{V1: With the advance of deep learning, object detection has achieved tremendous progress during the last decade, serving as a fundamental component in various multimedia applications such as video scene understanding, cross-modal retrieval, and Virtual Reality~(VR)~\cite{feichtenhofer2017detect,donahue2015long,karpathy2015deep,whelan2016elasticfusion,newcombe2011kinectfusion}. Despite their success, the generalization ability of object detectors still faces significant challenges, as numerous studies have demonstrated their substantial performance degradation on out-Of-distribution~(OOD) data~\cite{hendrycks2018benchmarking,hendrycks2019benchmarking,zhou2022domain}. For instance, an object detector trained for identifying pedestrians in urban environments performs well in clear weather but struggles in heavy rain and fog. To address the OOD issue, Single-Domain Generalized Object Detection~(S-DGOD) has emerged as a promising direction, which focuses on training object detectors within a single domain while improving their generalization across multiple unseen domains with the distribution shift~\citep{wu2022single,vidit2023clip,lee2024object}.}
With the advance of deep learning, object detection has achieved tremendous progress during the last decade, serving as a fundamental component in various multimedia applications such as video scene understanding, cross-modal retrieval, and Virtual Reality~(VR)~\cite{feichtenhofer2017detect,donahue2015long,karpathy2015deep,whelan2016elasticfusion,newcombe2011kinectfusion}. Despite their success, the generalization ability of object detectors still faces significant challenges, as numerous studies have demonstrated their substantial performance degradation on out-of-distribution~(OOD) data~\cite{hendrycks2018benchmarking,hendrycks2019benchmarking,zhou2022domain}. For instance, an object detector trained for identifying pedestrians in urban environments performs well in clear weather but struggles in heavy rain and fog. To address the OOD issue, Single-Domain Generalized Object Detection~(S-DGOD) has emerged as a promising direction, which focuses on training object detectors within a single domain while improving their generalization across multiple unseen domains with the distribution shift~\citep{wu2022single,vidit2023clip,lee2024object}.

Existing S-DGOD approaches can be categorized into two types: Vision-only paradigms and Vision-Language-Model-based~(VLM-based) paradigms. Vision-only paradigms primarily utilize multi-scale visual information to develop techniques such as data augmentation~\citep{lee2024object,xu2024physaug,danish2024improving}, contrastive learning~~\citep{lee2024object,danish2024improving}, and causal modeling~\citep{liu2024unbiased}, which reduce the distribution discrepancies between source and target domains within the vision-based feature space, as shown in Fig.~\ref{Figure 1}(a). PhysAug~\cite{xu2024physaug} simulates real-world domain shifts to augment visual training data, achieving state-of-the-art performance on S-DGOD tasks. In contrast, VLM-based methods leverage pre-trained knowledge from vision-language models to design implicit regularization, facilitating domain-invariant features in the vision-based latent space, as shown in Fig.~\ref{Figure 1}(b). CLIP the Gap~\cite{vidit2023clip} is the pioneer in using VLMs for S-DGOD, which utilizes scene prompts~(e.g. an image taken on a fog night) to perform semantic augmentations on vision-based features. Another work exploits scene prompts as implicit guidance to dynamically extract vision-based features in response to varying scenes~\cite{li2024prompt}. However, such approaches suffer from two limitations. First, their reliance on scene-level pre-trained knowledge is insufficient for learning the region-level feature invariance for accurate detection. Second, the inherent vision-language correlations, such as the correspondence between object regions and textual descriptions, are not explicitly leveraged to enhance the learning domain-invariant features.

% \textcolor{red}{In this paper, we propose a new cross-modal feature learning method, which enhances the generalization and discrimination of region-level features for S-DGOD tasks. Specifically, we introduce the Cross-modal and Region-aware Feature Interaction~(CRFI) mechanism. Different from prior VLM-based methods, CRFI explicitly models the inherent correlation between region-level visual content~(e.g., an image crop of a pedestrian) and textual descriptions~(e.g., a prompt of 'a pedestrian on a sunny day') within a joint image-text feature space. Building upon the correlation modeling and domain-invariant properties of VLMs, CRFI further enables dynamic interactions between visual features and textual features during training, thereby promoting both intra-modal and inter-modal regional feature invariance. Moreover, we introduce a new strategy called Cross-Domain Proposal Refining and Mixing~(CPRM) for better localization. CPRM simultaneously aligns and diversifies the position of region proposals across multiple domains, enhancing the localization ability of detectors in unseen domains.}

In this paper, we propose a new cross-modal feature learning method, which enhances the generalization and discrimination of region-level features for S-DGOD tasks. Specifically, we introduce the Cross-modal and Region-aware Feature Interaction~(CRFI) mechanism. Different from prior VLM-based methods, CRFI explicitly models the inherent correlation between region-level visual content~(e.g., an image crop of a pedestrian) and textual descriptions~(e.g., a prompt of 'a pedestrian on a sunny day') within a joint image-text feature space. Building upon the correlation modeling and domain-invariant properties of VLMs, CRFI further enables dynamic interactions between visual features and textual features during training, thereby promoting both intra-modal and inter-modal regional feature invariance. Moreover, we introduce a new strategy called Cross-domain Proposal Refining and Mixing~(CPRM) for better localization. CPRM simultaneously aligns and diversifies the position of region proposals across multiple domains, enhancing the localization ability of detectors in unseen domains.

% \textcolor{red}{We evaluate our proposed method on two popular S-DGOD benchmarks: Cityscapes-C and DWD. Experimental results demonstrate significant performance gains, improving by 8.8\%~mPC on Cityscapes-C and 7.9\%~mPC on DWD over the baseline method. Notably, our approach outperforms both existing vision-only and VLM-based methods, establishing new state-of-the-art performance. The main contributions of our work are as follows:
% \begin{itemize}
%     \item We propose a novel cross-modal feature learning method that enhances the generalization and discrimination of region-level features for S-DGOD tasks. This approach effectively bridges visual and textual modalities to enhance object detection across diverse unseen domains.
%     \item We introduce the Cross-modal and Region-aware Feature Interaction mechanism, which explicitly promotes dynamic interactions between region-level visual and textual features, ensuring robust intra- and inter-modal features.
%     \item We design the Cross-Domain Proposal Refining and Mixing strategy, which signs and diversifies region proposals to enhance the localization accuracy of detectors.
%     \item Our method achieves state-of-the-art performance on the Cityscapes-C and DWD benchmarks, surpassing previous VLM-based approaches with significant gains.
%     % \item Our method achieves state-of-the-art performance on both the Cityscapes-C and DWD benchmarks, surpassing previous vision-language-based approaches with substantial improvements.
% \end{itemize}}

We evaluate our proposed method on two popular S-DGOD benchmarks: Cityscapes-C and DWD. Experimental results demonstrate significant performance gains, improving by 8.8\%~mPC on Cityscapes-C and 7.9\%~mPC on DWD over the baseline method. Notably, our approach outperforms both existing Vision-only and VLM-based methods, establishing new state-of-the-art performance. The main contributions of our work are as follows:
\begin{itemize}
    \item We propose a novel cross-modal feature learning method that enhances the generalization and discrimination of region-level features for S-DGOD tasks. This approach effectively bridges visual and textual modalities to enhance object detection across diverse unseen domains.
    \item We introduce the Cross-modal and Region-aware Feature Interaction mechanism, which explicitly promotes dynamic interactions between region-level visual and textual features, ensuring robust intra- and inter-modal features.
    \item We design the Cross-domain Proposal Refining and Mixing strategy, which signs and diversifies region proposals to enhance the localization accuracy of detectors.
    % \item Our method achieves state-of-the-art performance on the Cityscapes-C and DWD benchmarks, surpassing previous VLM-based approaches with significant gains.
    \item Our method achieves state-of-the-art performance on the DWD and Cityscapes-C benchmarks, surpassing previous approaches with significant gains.
\end{itemize}

\section{RELATED WORK}
\subsection{Single Domain Generalized Object Detection}
In recent years, Single Domain Generalized Object Detection (S-DGOD) has emerged as a hot research direction to address the robustness challenges in real-world vision systems. The community has witnessed promising progress through diverse methodological avenues. Vision-only methods usually employ augmentation strategies~\citep{zhou2022domain,shorten2019survey,zhao2020maximum,rebuffi2021data,modas2022prime,vaish2024fourier,zhang2017mixup,devries2017improved,hendrycks2019augmix,cubuk2020randaugment} to improve model robustness against domain shifts. OA-DG ~\cite{lee2024object} utilizes object-aware data augmentation with contrastive loss, effectively decoupling ROI features from irrelevant background regions. DivAlign~\citep{danish2024improving} advances this paradigm by harmonizing domain diversity expansion with task-aware loss calibration. Physics-inspired frameworks like PhysAug~\cite {xu2024physaug} further expand the scope by leveraging the material reflectance principle. 
Moreover, significant progress has also been made in Vision-only methods' model architecture design, particularly through innovations like CDSD~\cite{wu2022single} and G-NAS~\cite{wu2024g}.
Cyclic self-decoupling (CDSD) demonstrates notable efficacy against texture-level domain shifts through reconstruction-guided feature disentanglement, and gradient-aware neural search (G-NAS) achieves model-adaptive architecture for style scenarios by adaptively optimizing network topology. 
Concurrently, Vision-Language Model (VLM)-based methods leverage semantic augmentation techniques to enhance cross-modal feature diversity. By employing textual and visual knowledge through CLIP-driven prompting mechanisms~\cite{vidit2023clip,fahes2023poda}, these approaches synthesize domain-adaptive representations, and environmental text prompts are encoded to perturb image embeddings, thereby simulating diverse weather conditions and lighting variations.
The above methods either design various constraints based on visual knowledge or utilize scene-level textual information to guide domain-invariant learning in image feature space. In contrast, our work focuses on the interactions between visual features and textual features, fully using detailed textual information to separate domain differences and obtain cross-modal domain-invariant representations for detection.

% \begin{figure*}[!ht]
%     \centering
%     \includegraphics[width=\linewidth]{figure2.png}
%     \caption{This figure illustrates our Multi-Level Text-Image Invariance Learning framework. The Scene-level Proposals Alignment module enables the model to gain a wider scene context to support the extraction of domain-invariant features at the global level. The Dual-Level Text-Image Alignment module extracts object and background-level invariant features through text-image interaction. The overall optimization is guided by $loss_{roi}$, $loss_{slp}$, and $loss_{ti}$, improving domain generalization.}
%     \label{Figure 2}
% \end{figure*}

\subsection{Vision-Language Models for Domain Generalization}
Joint vision-language representation learning has achieved remarkable progress through contrastive text-image alignment~\cite {radford2021learning, desai2021virtex}, which has unlocked novel approaches for domain generalization through their inherent text-image alignment capabilities. Works like DetCLIP~\citep{yao2022detclip} and DenseCLIP~\citep{rao2022denseclip} exploit CLIP's semantic alignment capabilities to refine region-level image features via textual prompts, while CLIPGap~\citep{vidit2023clip} and CLIP-Cluster~\citep{shen2023clip} mitigate domain gaps by aligning task-specific features with CLIP’s global embeddings. These approaches primarily focus on leveraging correlational links between text and image features to enhance open-vocabulary detection or domain-invariant classification. Text prompt tuning methods~\cite {zhou2022learning, zhou2022conditional} address this limitation by learning continuous prompt vectors to adapt CLIP embeddings to target tasks dynamically. Nevertheless, current adaptive methods are mainly focused on optimizing statistical correlations, but they do not explicitly establish mechanisms for the interaction between prompt semantics and cross-domain invariant image features. In comparison, our work introduces a text-image invariance learning paradigm that establishes robust links between textual prompts and invariant image features.

\section{METHODOLOGY}
\subsection{Problem Setting}
Single-domain Generalization for Object Detection (S-DGOD) seeks to train an object detector $F$ on a single labeled source domain $D_s=\{( x_{i}^{s},y_{i}^{s} ) \}_{i=1}^{N_s}$, where $x_i^s$ represents input images and $y_i^s$ denotes corresponding labels, and enable robust performance across multiple unseen target domains $M=\{ D_{t}^{1},D_{t}^{2},\cdots ,D_{t}^{k} \}$, with each $D_{t}^{n}=\{( x_{i}^{t} ) \}_{i=1}^{N_t}$ containing unlabeled images~\citep{wu2022single}. Here, $N_s$ and $N_t$ denote the sizes of the source and target datasets, respectively. 

The primary challenge arises from the distribution shift between $D_s$ and the target domains in $M$, driven by variations in scene composition, object appearance, lighting, weather, and image quality~\citep{wang2021robust}. This shift degrades the performance of the detector $F$ trained only on $D_s$ when applied to $D_t^n \in M$. Thus, S-DGOD requires detector $F$ to learn domain-invariant representations for robust generalization to unseen target domains without their data during training. Formally, the objective is to maximize the detection performance across the target domains:
\begin{equation}
\max_F \mathbb{E}_{D_t^n \sim M} \left[ \text{mAP}(F, D_t^n) \right],
\end{equation}
where $\text{mAP}(F, D_t^n)$ represents the mean Average Precision, averaging the precision-recall performance across all object classes for detector $F$ on target domain $D_t^n$.

\subsection{Overview}
Our proposed cross-modal feature learning method for S-DGOD tasks integrates two pivotal components: the Cross-modal and Region-aware Feature Interaction (CRFI) mechanism and the Cross-domain Proposal Refining and Mixing (CPRM) strategy.
% Our method consists of Cross-modal and Region-aware Feature Interaction~(CRFI) and Cross-domain Proposal Refining and Mixing~(CPRM). 
CRFI models the interactions among multiple regions by leveraging cross-domain image information and cross-modal image-text relationships. Through these interactions, CRFI dynamically clusters regions of the same category across different modalities and domains while enforcing greater feature separation between regions of different categories, ensuring more discriminative and domain-invariant representations. 
To further enhance object localization performance in domain-specific environments, CPRM, which is inherently complementary to CRFI, refines region proposal positions across clean and augmented domains and mixes them to capture diverse structural and semantic variations, enabling the detector to accurately and consistently localize objects across a wide range of unseen scenarios.
% CRFI extracts domain-invariant features by interacting with multi-modal representations across different regions. CPRM provides proposal spatial guidance, ensuring consistent localization of regions of interest across different domains, reducing ambiguity in noisy conditions.
In the following sections, we provide a detailed explanation of each component of our framework.
\begin{figure*}[!ht]
    \centering
    \includegraphics[width=\linewidth]{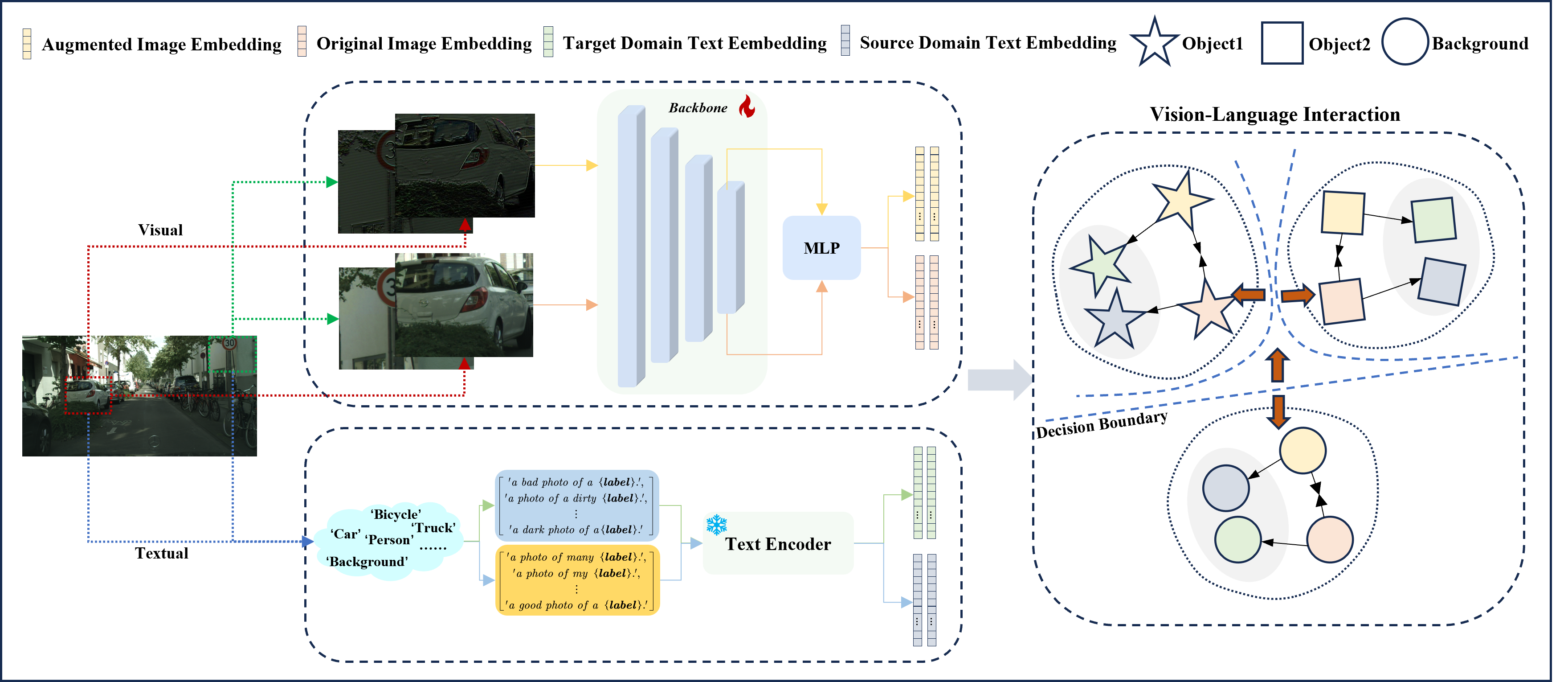}
    \caption{
    This figure illustrates our CRFI, which extracts multi-region invariant features through text-image interaction.
    % Cross-Domain Proposal Refining and Mixing~(CPRM) enables the model to focus on shared domain-invariant features across different domains.
    }
    \label{Figure 2}
\end{figure*}

% \begin{figure}[!ht]
%     \centering
%     \includegraphics[width=\linewidth]{rrl_fig.png}
%     \caption{
%     This figure illustrates our CPRM, which enables RPN to increasingly localize domain-invariant proposals across domains as training proceeds.}
%     \label{CPRM_fig}
% \end{figure}

\subsection{Cross-modal and Region-aware Feature Interaction}
The illustration of Cross-modal and Region-aware Feature Interaction is shown in Fig.~\ref{Figure 2}.
To introduce domain discrepancies between the image and text domains under complex environments, we first employ data augmentation techniques to simulate domain distributions in noisy conditions. 
% Unless otherwise stated, CRFI in the following sections adopts our proposed data augmentation strategy by default, which serves as the standard configuration for all related experiments. 
Additionally, the prompts template $P$ is adapted based on whether the regions are clean or augmented, denoted as $P_{ori}$ and $P_{aug}$, respectively.

To achieve text-image interactions at different regions, including both object and background regions, we introduce a selection strategy that ensures a balanced mapping between textual and visual features, as is shown in Algorithm~\ref {selection_alg}. Given an input image with multiple object instances across different categories, we select representative object regions per unique category and a background region. The background region is explicitly chosen from areas without any foreground objects, ensuring that it represents background context rather than specific object information. In parallel, we convert the corresponding object labels and background descriptions into prompts. 

% \begin{algorithm}[H]
% \caption{Multi-Region Text-Image Selection}
% \label{selection_alg}
% \begin{algorithmic}[1]
% \Require Batch of Images: \(\mathcal{I} = \{I_1, I_2, \dots, I_N\} \in D_S\). \par
% Prompts Template \(P\)
% \Ensure Aggregated sets \(O, B, C_O, C_B\)
% \State \(O, B, C_O, C_B \gets \emptyset\)
% \ForAll{\(I \in \mathcal{I}\)}
%     \ForAll{\(c \in U(I)\)}   \Comment{\(U(I)\): unique categories in \(I\)}
%         \State \(o^*_c \gets \text{select}(\{o \in I : \text{category}(o)=c\})\)
%         \State \(O \gets O \cup \{o^*_c\}\), \(C_O \gets C_O \cup \{c\}\)  
%     \EndFor
%     \State \(B \gets B \cup \text{selectBackground}(I)\), \(C_B \gets C_B \cup \{\text{"background"}\}\)
% \EndFor
% \State \Return \(O, B, P(C_O), P(C_B)\)
% \end{algorithmic}
% \end{algorithm}

\begin{algorithm}[H]
\caption{Multi-Region Text-Image Selection}
\label{selection_alg}
\begin{algorithmic}[1]
% \Require Batch of Images: \(\mathcal{I} = \{I_1, I_2, \dots, I_N\} \in D_S\). \par
\Require Batch of Images: \(\mathcal{I} = \{( x_{i}^{s},y_{i}^{s} ) \}_{i=1}^{N}
 \in D_S\). \par
Prompts Template \(P\)
\Ensure Aggregated sets \(O, B, C_{obj}, C_{bg}\)
\State Initialize \(O, B, C_{obj}, C_{bg}\) as empty sets
\ForAll{\(I \in \mathcal{I}\)}
    \ForAll{\(c \in U(I)\)}   \Comment{\(U(I)\): unique categories in \(I\)}
        \State \(o^*_c \gets \text{select}(\{o \in I : \text{category}(o)=c\})\)
        \State \(O \gets O \cup \{o^*_c\}\), \(C_{obj} \gets C_O \cup \{c\}\)  
    \EndFor
    \State \(B \gets B \cup \text{selectBackground}(I)\)
    \State \(C_{bg} \gets C_{bg} \cup \{\text{"background"}\}\)
\EndFor
\State \Return \(O, B, P(C_{obj}), P(C_{bg})\)
\end{algorithmic}
\end{algorithm}

% \begin{algorithm}[H]
% \caption{Multi-Region Text-Image Selection}
% \label{selection_alg}
% \begin{algorithmic}[1]
% \Require Batch of Images: \(\mathcal{I} = \{I_1, I_2, \dots, I_N\} \in D_S\) \Comment{Labeled source domain images}
% \Require Prompts Template \(P\)  \Comment{Category-specific text prompts}
% \Ensure Aggregated sets \(O, B, C_O, C_B\)  
% \State \(O, B, C_O, C_B \gets \emptyset\)
% \ForAll{\(I \in \mathcal{I}\)}
%     \ForAll{\(c \in U(I)\)} \Comment{\(U(I)\): Unique categories in \(I\)}
%         \State \(o^*_c \gets \text{select}(\{o \in I : \text{category}(o)=c\})\)  \Comment{Selects a representative object}
%         \State \(O \gets O \cup \{o^*_c\}\), \(C_O \gets C_O \cup \{c\}\)  
%     \EndFor
%     \State \(B \gets B \cup \text{selectBackground}(I)\) \Comment{Extracts background regions}
%     \State \(C_B \gets C_B \cup \{\text{"background"}\}\) 
% \EndFor
% \end{algorithmic}
% \end{algorithm}

\begin{figure}[!ht]
    \centering
    \includegraphics[width=\linewidth]{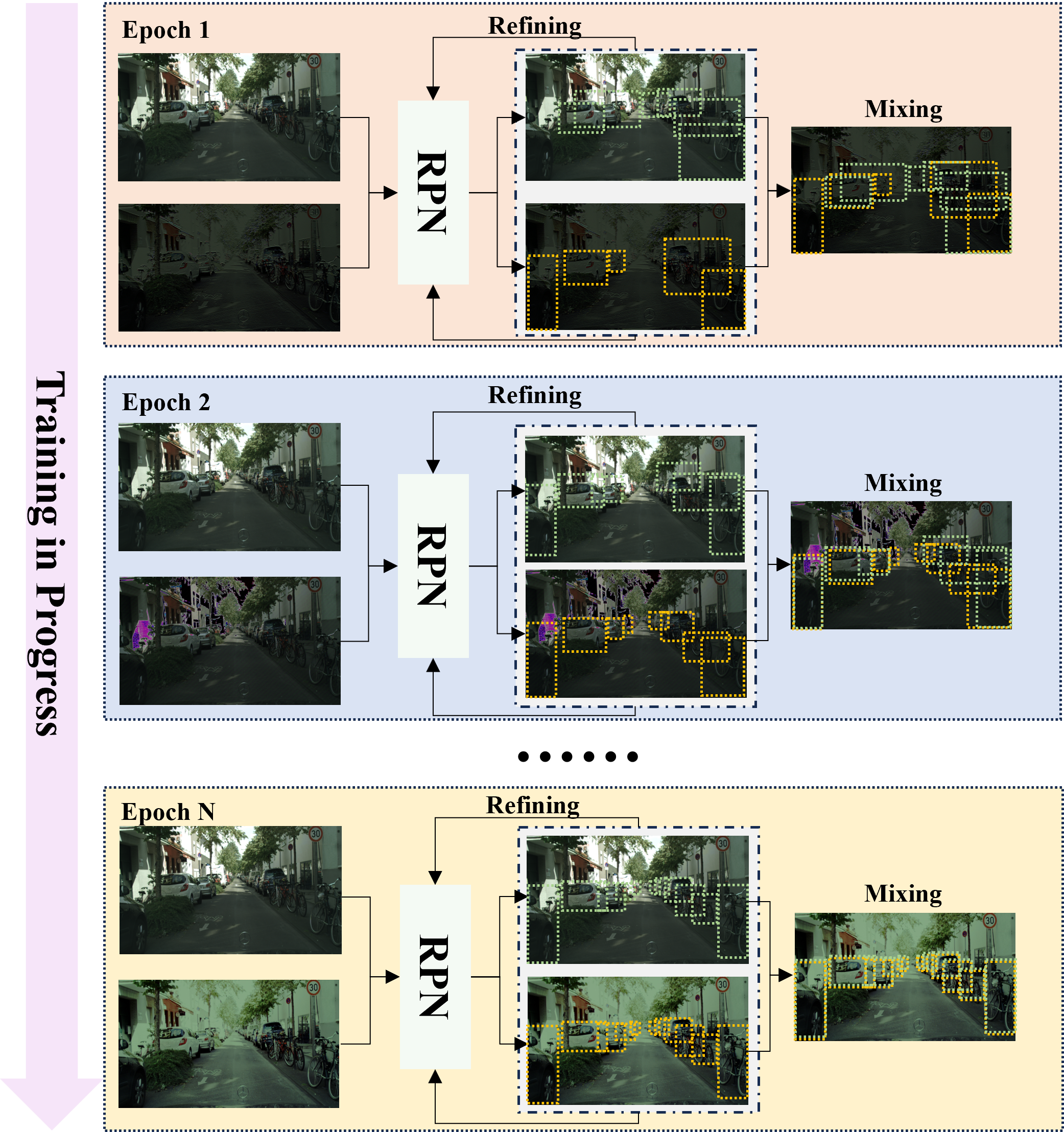}
    \caption{
    This figure illustrates our CPRM, which enables RPN to increasingly localize domain-invariant proposals across domains as training proceeds.}
    \label{CPRM_fig}
\end{figure}

Through the multi-region text-image selection strategy, the regions we get are stacked within each image and aggregated across the batch. The regions $O$, $B$ are then processed by the backbone to extract different regions' features. Additionally, we generate text features from textual prompts $P(C_O)$, $P(C_B)$.
We express the relevant regions' features obtained from both augmented and clean images as follows:
\begin{equation}
\begin{array}{ll}
\begin{cases}
	% t_{obj}=g\left( P(C_O)_{ori} \right) ,&		
        T^{s}_{\text{obj}}=g\left( P_{ori}(C_{obj}) \right) ,&		
    T^{t}_{\text{obj}}=g\left( P_{aug}(C_{obj}) \right)\\
	I^{s}_{\text{obj}}=f\left( O_{ori} \right) ,&		
    I^{t}_{\text{obj}}=f\left( O_{aug} \right)
\end{cases}
\\
\begin{cases}
	T^{s}_{\text{bg}}=g\left( P_{ori}(C_{bg}) \right) ,&		
    T^{t}_{\text{bg}}=g\left( P_{aug}(C_{bg}) \right)\\
	I^{s}_{\text{bg}}=f\left( B_{ori} \right) ,&		
    I^{t}_{\text{bg}}=f\left( B_{aug} \right)
\end{cases}
\end{array}
\label{Eq.(3)}
\end{equation}

In Eq.~\ref{Eq.(3)}, \(g\) denotes the frozen CLIP text encoder, while \(f\) represents the backbone of the object detector. The prompts of the different regions before and after augmentation are denoted as \( P_{ori}(C_{obj}), P_{aug}(C_{obj}), P_{ori}(C_{bg}), \) and \( P_{aug}(C_{bg}) \), respectively, with their corresponding extracted text features given by \( T^{s}_{\text{obj}}, T^{t}_{\text{obj}}, T^{s}_{\text{bg}}, \) and \( T^{t}_{\text{bg}} \). Similarly, the regions' images before and after augmentation are represented as \( O_{ori}, O_{aug}, B_{ori}, \) and \( B_{aug} \), with their extracted image features denoted as \( I^{s}_{\text{obj}}, I^{t}_{\text{obj}}, I^{s}_{\text{bg}}, \) and \( I^{t}_{\text{bg}} \).

To achieve pairwise interaction between multi-region textual semantics and image features before and after augmentation, we design a series of contrastive losses based on \( T^{s}_{\text{obj}}, T^{t}_{\text{obj}}, T^{s}_{\text{bg}}, \) and \( T^{t}_{\text{bg}} \). as well as \( I^{s}_{\text{obj}}, I^{t}_{\text{obj}}, I^{s}_{\text{bg}}, \) and \( I^{t}_{\text{bg}} \). This interaction facilitates the extraction of domain-invariant features across different regions, ensuring that the model's representations generalize effectively to new, unseen domains.

As illustrated in Fig.~\ref{Figure 2}, the CRFI models the relationships between different modalities, focusing on learning both inter-modal and intra-modal regional invariance. This knowledge interaction not only aligns textual and visual features at the object region but also integrates contextual information from the background. To achieve multi-region interaction, we employ InfoNCE to compute the text-image losses. Identity labels are constructed for each feature embedding to ensure that it is most similar to its corresponding counterpart while contrasting against negative samples. This strategy encourages clustering of similar regions across text-image pairs, reducing cross-modal and image-domain discrepancies while ensuring that non-similar regions maintain a clear separation.
Through multi-region interactions, the model develops a more comprehensive understanding of the scene, allowing it to better disentangle domain-specific variations and enhance feature robustness.
% This loss formulation ensures proper alignment of object and background-level text and image features, enabling the model to effectively extract consistent, domain-invariant features that improve performance in unseen target domains.
The Cross-modal and Region-aware Feature Interaction loss, denoted as \(\mathcal{L}_{\text{CRFI}}\), is formulated as follows:
\begin{equation}
\mathcal{L}_{\text{CRFI}}=\frac{1}{2}[\mathcal{L}_{\{i_{\text{ori}},i_{\text{aug}}\}}+\frac{1}{2}(\mathcal{L}_{\{t_{\text{obj}},i_{\text{obj}}\}}+\mathcal{L}_{\{t_{\text{bg}},i_{\text{bg}}\}}) ]
\label{Eq.(4)}
\end{equation}

The $\mathcal{L}_{\{i_{\text{aug}},i_{\text{ori}}\}}, \mathcal{L}_{\{t_{\text{obj}},i_{\text{obj}}\}}, \mathcal{L}_{\{t_{\text{bg}},i_{\text{bg}}\}}$ in Eq.~\ref{Eq.(4)} are given below. In the following equation, $\langle, \rangle$ represents the concatenation of two tensors, $\Gamma$ represents the InfoNCE loss.
\begin{equation}
\begin{aligned}
    \mathcal{L}_{\{i_{\text{aug}},i_{\text{ori}}\}} &= \text{$\Gamma$} \left( \langle I^{s}_{\text{obj}}, I^{s}_{\text{bg}} \rangle, \langle I^{t}_{\text{obj}}, I^{t}_{\text{bg}} \rangle \right)
\end{aligned}
\label{Eq.(5)}
\end{equation}

\begin{equation}
\begin{aligned}
\mathcal{L}_{\{{t_{\text{obj}}},i_{\text{obj}}\}} &= \frac{1}{2} \left( \text{$\Gamma$} \left( I^{s}_{\text{obj}}, T^{s}_{\text{obj}} \right) + \text{$\Gamma$} \left( I^{t}_{\text{obj}}, T^{t}_{\text{obj}} \right) \right)
\end{aligned}
\label{Eq.(6)}
\end{equation}

\begin{equation}
\begin{aligned}
\mathcal{L}_{\{{t_{\text{bg}}},i_{\text{bg}}\}} &= \frac{1}{2} \left( \text{$\Gamma$} \left( I^{s}_{\text{bg}}, T^{s}_{\text{bg}} \right) + \text{$\Gamma$} \left( I^{t}_{\text{bg}}, T^{t}_{\text{bg}} \right) \right)
\end{aligned}
\label{Eq.(7)}
\end{equation}

\begin{table*}[t]
\centering
\fontsize{9}{12}\selectfont % Set font size to 10pt with 12pt line spacing
\setlength{\tabcolsep}{0.8mm} % Adjust column width to fit within the page
\caption{Comparison of our method with state-of-the-art approaches on Cityscapes-C. The mPC metric is defined in section 4.1.2. Bold numbers indicate the best results, while underlined values denote the second-best performance.}
\label{Table 1}
\begin{tabular}{c|c|ccccccccccccccc|c}
\hline
\multirow{2}{*}{Method} & \multirow{2}{*}{Clean} & \multicolumn{3}{c}{Noise} & \multicolumn{4}{c}{Blur} & \multicolumn{4}{c}{Weather} & \multicolumn{4}{c|}{Digital} & \multirow{2}{*}{mPC} \\ \cline{3-17}
& & Gauss & Shot & Impulse & Defocus & Glass & Motion & Zoom & Snow & Frost & Fog & Bright & Contrast & Elastic & Pixel & JPEG \\ \hline
baseline & 42.2 & 0.5 & 1.1 & 1.1 & 17.2 & 16.5 & 18.3 & 2.1 & 2.2 & 12.3 & 29.8 & 32.0 & 24.1 & \underline{40.1} & 18.7 & 15.1 & 15.4 \\ \hline
% \multicolumn{18}{c}{\textbf{Vision-Only Methods}} \\ \hline
Cutout & 42.5 & 0.6 & 1.2 & 1.2 & 17.8 & 15.9 & 18.9 & 2.0 & 2.5 & 13.6 & 29.8 & 32.3 & 24.6 & \underline{40.1} & 18.9 & 15.6 & 15.7 \\
PhotoDistort & 42.7 & 1.6 & 2.7 & 1.9 & 17.9 & 14.1 & 18.7 & 2.0 & 2.4 & 16.5 & 36.0 & 39.1 & 27.1 & 39.7 & 18.0 & 16.4 & 16.9 \\
AutoAug & 42.4 & 0.9 & 1.6 & 0.9 & 16.8 & 14.4 & 18.9 & 2.0 & 1.9 & 16.0 & 32.9 & 35.2 & 26.3 & 39.4 & 17.9 & 11.6 & 15.8 \\
AugMix & 39.5 & 5.0 & 6.8 & 5.1 & 18.3 & 18.1 & 19.3 & \textbf{6.2} & 5.0 & 20.5 & 31.2 & 33.7 & 25.6 & 37.4 & 20.3 & 19.6 & 18.1 \\
StylizedAug & 36.3 & 4.8 & 6.8 & 4.3 & 19.5 & 18.7 & 18.5 & 2.7 & 3.5 & 17.0 & 30.5 & 31.9 & 22.7 & 33.9 & 22.6 & 20.8 & 17.2 \\
OA-Mix & 42.7 & 7.2 & 9.6 & 7.7 & 22.8 & 18.8 & 21.9 & \underline{5.4} & 5.2 & 23.6 & 37.3 & 38.7 & 31.9 & \textbf{40.2} & 22.2 & 20.2 & 20.8 \\ 
SupCon & \underline{43.2} & 7.0 & 9.5 & 7.4 & 22.6 & 20.2 & 22.3 & 4.3 & 5.3 & 23.0 & 37.3 & 38.9 & 31.6 & 40.1 & 24.0 & 20.1 & 20.9 \\
FSCE & 43.1 & 7.4 & 10.2 & 8.2 & 23.3 & 20.3 & 21.5 & 4.8 & 5.6 & 23.6 & 37.1 & 38.0 & 31.9 & 40.0 & 23.2 & 20.4 & 21.0 \\
OA-DG & \textbf{43.4} & 8.2 & 10.6 & 8.4 & 24.6 & \textbf{20.5} & 22.3 & 4.8 & 6.1 & \textbf{25.0} & \underline{38.4} & \underline{39.7} & \textbf{32.8} & \textbf{40.2} &23.8 & \textbf{22.0} & 21.8 \\
PhysAug & 42.6 & \underline{14.3} & \underline{17.0} & \underline{11.9} & \underline{25.6} & 19.1 & \textbf{25.5} & 3.9 & \underline{8.6} & 21.3 & 35.3 & 39.5 & 27.5 & 39.1 & \textbf{28.9} & 19.9 & \underline{22.6} \\ \hline

% \multicolumn{18}{c}{\textbf{VLM-based Method}} \\ \hline
Ours & 42.2 & \textbf{17.4} & \textbf{20.5} & \textbf{15.6} & \textbf{26.6} & \underline{19.7} & \underline{25.3} & 4.4 & \textbf{10.0} & \underline{23.3} & \textbf{39.3} & \textbf{40.0} & \underline{31.9} & 39.4 & \underline{28.7} & \underline{21.2} & \textbf{24.2} \\ \hline
% \multicolumn{18}{c}{\textbf{Ours}} \\ \hline
\end{tabular}
\end{table*}

% \begin{table}[]
% \caption{Average Recall (AR) results for maxDets=1000 under different object sizes on Cityscapes.}
% \label{CPRM}
% \centering
% \renewcommand{\arraystretch}{1.3}
% \fontsize{10pt}{12pt}\selectfont
% \begin{tabular}{l|c|c|c|c}
% \hline
% Method   & all  & small & medium & large \\ \hline
% Baseline & 51.4 & 26.1  & 52.5   & 71.2  \\ \hline
% CRFI    & 50.3 & 26.3  & 50.7   & 70.8  \\ \hline
% \end{tabular}
% \end{table}

% \begin{figure}[!ht]
%     \centering
%     \includegraphics[width=\linewidth]{ar_comparison.png}
%     \caption{
%     Average Recall~(AR) results for maxDets=1000 under different object sizes on the Cityscapes validation set.}
%     \label{CPRM}
% \end{figure}

% \begin{figure}[!ht]
%     \centering
%     \includegraphics[width=\linewidth]{rrl_fig.png}
%     \caption{
%     This figure illustrates our CPRM, which enables RPN to increasingly localize domain-invariant proposals across domains as training proceeds.}
%     \label{CPRM_fig}
% \end{figure}

\subsection{Cross-domain Proposal Refining and Mixing}
% The Cross-Modal and Region-Aware Feature Interaction~(CRFI) ensures interactions between multi-region visual and text features before and after augmentation. 
% But we found that the CRFI also led to a certain degree of degradation in the model’s localization ability under the clean validation set, as evidenced by the decrease in Average Recall, as shown in Fig.~\ref{CPRM}. 
To further enhance the model's localization ability, we introduce a simple but effective strategy called Cross-domain Proposal Refining and Mixing~(CPRM), which aims to strengthen the model's ability to localize domain-invariant features across diverse domains. The illustration of Cross-domain Proposal Refining and Mixing~(CPRM) is shown in Fig.~\ref{CPRM_fig}.

To achieve regional localization refinement, the Region Proposal Network (RPN) is used to extract region-of-interest features from both clean and augmented images. Throughout training, the model continuously refines the RPN’s localization capability across both clean and augmented domains, effectively preventing the loss of regions of interest caused by domain shifts and preserving localization consistency across domains. The corresponding $\mathcal{L}_{\text{CPRM}}$~($\mathcal{L}_{\text{refine}}$) is defined as follows, which serves as a replacement for the RPN-related loss:
\begin{equation}
% loss_{refine}=\frac{1}{2}(loss_{rpn_{aug}}+loss_{rpn_{ori}})
\mathcal{L}_{\text{CPRM}} =\mathcal{L}_{\text{refine}} = \frac{1}{2} \left( \mathcal{L}_{\text{rpn}}^{\text{aug}} + \mathcal{L}_{\text{rpn}}^{\text{ori}} \right)
\label{Eq.(9)}
\end{equation}
In addition, CPRM mixes the proposals generated from clean and augmented images, allowing the model to learn consistent region localization across domains by capturing broader structural and semantic variations.
The following equation can express the concept of proposal mixing:
\begin{equation}
R_{\text{multi-region}}=<R_{\text{region}}^{\text{ori}},R_{\text{region}}^{\text{aug}}>
\label{Eq.(8)}
\end{equation}
% This process captures broader structural and semantic variations caused by domain shifts, ensuring that the model learns consistent region localization across different domains. 
% The following equation can express the concept of this module:
% \begin{equation}
% R_{multi-region}=<R_{region_{ori}},R_{region_{aug}}>
% \label{Eq.(8)}
% \end{equation}
The $R_{\text{multi-region}}, R_{\text{region}}^{\text{ori}}$, and $R_{\text{region}}^{\text{aug}}$ in Eq.~\ref{Eq.(8)} represent the ROIs produced by CPRM, the ROIs from the clean image, and the ROIs from the augmented image, respectively. The $R_{multi-region}$ will be forwarded to the next training stage.
Mixing proposals from clean and augmented views encourages the model to focus on domain-invariant cues, thus enhancing its generalization to unseen target domains.
% The Cross-Domain Proposal Refining and Mixing loss is expressed as follows, which serves as a replacement for the RPN-related loss:
% \begin{equation}
% loss_{refine}=\frac{1}{2}(loss_{rpn_{aug}}+loss_{rpn_{ori}})
% \label{Eq.(9)}
% \end{equation}

In general, CPRM refines the localization of regions across clean and augmented images, ensuring consistent region positioning. Meanwhile, the CRFI ensures consistent semantic relationships between text and image regions, making the features domain-invariant. 
These techniques complement each other by aligning semantic features across modals and domains and refining region localization, improving both feature extraction and object detection robustness across diverse domains.

The overall optimization is driven by ROI loss~$\mathcal{L}_{\text{roi}}$, Cross-modal and Region-aware Feature Interaction loss~$\mathcal{L}_{\text{CRFI}}$, and Cross-domain Proposal Refining and Mixing loss~$\mathcal{L}_{\text{CPRM}}$, enhancing domain generalization. The complete loss formulation $\mathcal{L}$ is defined as follows, where $\alpha$ is a hyperparameter.
\begin{equation}
\mathcal{L} = \mathcal{L}_{\text{roi}}+\mathcal{L}_{\text{CPRM}}+\alpha*\mathcal{L}_{\text{CRFI}}
\label{Eq.(10)}
\end{equation}

\begin{table*}[t!]
\centering
\fontsize{10}{12}\selectfont
\caption{Comparison with state-of-the-art Vision-only and VLM-based methods on the Diverse Weather Dataset~(DWD). The mPC metric is defined in section 4.1.2. Bold numbers represent the highest performance in each column, and underlined numbers indicate the second-highest rank.}
\label{Table 2}
\begin{tabular}{c|c|cccc|c}
\hline
Method & Daytime Sunny &Night Sunny&Dusk Rainy&Night Rainy&Daytime Foggy & mPC \\ \hline
Baseline  & 50.4 & 37.5 & 29.2 & 14.6 & 33.1 & 30.2 \\ \hline
\multicolumn{7}{c}{\textbf{Vision-only Methods}} \\ \hline
IBN-Net  & 49.7 & 32.1 & 26.1 & 14.3 & 29.6 & 25.5 \\
SW & 50.6 & 33.4 & 26.3 & 13.7 & 30.8 & 26.1 \\
IterNorm & 43.9 & 29.6 & 22.8 & 12.6 & 28.4 & 23.4 \\
ISW & 51.3 & 33.2 & 25.9 & 14.1 & 31.8 & 26.3 \\
SHADE & -- & 33.9 & 29.5 & 16.8 & 33.4 & 28.4 \\
CDSD & 56.1 & 36.6 & 28.2 & 16.6 & 33.5 & 28.7 \\
SRCD & -- & 36.7 & 28.8 & 17.0 & 35.9 & 29.6 \\
OA-Mix & 56.4 & 38.6 & 33.8 & 14.8 & 38.1 & 31.3 \\
OA-DG & 55.8 & 38.0 & 33.9 & 16.8 & 38.3 & 31.8 \\
Div & 50.6 & 39.4 & 37.0 & 22.0 & 35.6 & 33.5 \\
DivAlign & 52.8 &42.5 & 38.1 & \textbf{24.1} & 37.2 & 35.5 \\
UFR & 58.6 & 40.8 & 33.2 & 19.2 & 39.6 & 33.2 \\
PhysAug & \underline{60.2} & \underline{44.9} & \underline{41.2} & 23.1 & 40.8 & \underline{37.5} \\ \hline 
\multicolumn{7}{c}{\textbf{VLM-based Methods}} \\ \hline
CLIP-Gap & 51.3 & 36.9 & 32.3 & 18.7 & 38.5 & 31.6 \\
Li's & 53.6 & 38.5 & 33.7 & 19.2 & 39.1 & 32.6 \\
VLTDet & \textbf{60.5} & 44.6 & 38.4 & 22.1 & \textbf{42.3} & 36.9 \\
Ours & 59.8 & \textbf{45.0} & \textbf{41.3} & \underline{23.8} &  \underline{42.1} & \textbf{38.1} \\ \hline
\end{tabular}
\end{table*}

\section{EXPERIMENTS}
In this section, we present the benchmarks and experimental setup used to evaluate our approach. We will showcase the results across different single-domain generalization object detection benchmarks and provide detailed ablation studies.
\subsection{Setting Up}

\subsubsection{Datasets} Our evaluations are conducted on Cityscapes-C~\cite {hendrycks2019benchmarking} and Diverse Weather Dataset~(DWD)~\citep{wu2022single}, benchmarks designed to systematically assess the robustness of object detection models under diverse image degradations. Cityscapes-C is generated by applying 15 different corruption types, each with five severity levels, to the validation set of Cityscapes~\cite {michaelis2019benchmarking}. These corruptions are grouped into four major categories: Noise, Blur, Digital, and Weather-related distortions. The dataset provides a controlled evaluation setting for measuring a model’s ability to generalize across real-world degradation factors. Following S-DGOD protocols, we use the training set of Cityscapes as the source domain, while corrupted versions of the validation set serve as unseen target domains for evaluation.
In addition to synthetic corruptions, we choose the DWD, a real-world urban-scene detection dataset collected from multiple autonomous driving benchmarks~\citep{yu2020bdd100k,sakaridis2018semantic,hassaballah2020vehicle}. 
By evaluating our approach on both Cityscapes-C and DWD, we provide a comprehensive assessment of its effectiveness across both synthetic corruptions and real-world weather variations, ensuring robustness to diverse domain shifts.

\subsubsection{Evaluation Metrics}
To assess the performance of our method on the clean domain, we use mean average precision (mAP), following the guidelines established by Michaelis et al.\citep{michaelis2019benchmarking}. In addition, we introduce the mean performance under corruption (mPC) metric, as defined by Lee et al.\citep{lee2024object}, to evaluate performance across corrupted domains. The mPC provides an aggregate measure of mAP performance over various corruptions and severity levels, and is calculated as follows:
\begin{equation}
mPC = \frac{1}{N_C} \sum_{C=1}^{N_C} \left( \frac{1}{N_S} \sum_{S=1}^{N_S} P_{C,S} \right),
\label{Eq.(11)}
\end{equation}
where $P_{C,S}$ refers to the mAP of the target domain under corruption $C$ at severity level $S$. The parameters $N_C$ and $N_S$ represent the number of corruption types and severity levels, respectively. 
For Cityscapes-C, $N_C$ is 15 and $N_S$ is 5, while for DWD dataset, $N_C$ is 4 and $N_S$ is 1

\subsubsection{Implementation}
In the experiments, our configurations all align with those in CDSD, OA-DG, and Physaug.
In CRFI, we apply random color and frequency transformations for data augmentation, as detailed in the supplementary material.
For experiments on the Cityscapes-C dataset, we utilize Faster R-CNN~\citep{ren2015faster} with a ResNet50~\citep{he2016deep} backbone and incorporate feature pyramid networks (FPN)~\citep{lin2017feature}. The learning rate is set to $0.01$, and we use a batch size of $8$. The optimizer is SGD, with a momentum of $0.9$ and a weight decay of $0.0001$. 
For the DWD dataset, we use Faster R-CNN~\citep{ren2015faster} with a ResNet101~\citep{he2016deep} backbone as the model. We use the SGD optimizer with a momentum of $0.9$ and a weight decay of $0.0001$. The learning rate is set to $0.001$, with a batch size of $2$.
In our implementation, the CLIP text encoder employs a ResNet101 backbone, and we set $\alpha$ values to $0.01$ and $0.03$ for the respective experiments. 
All other configurations align with those in CDSD~\citep{wu2022single}, OA-DG~\citep{lee2024object}, and Physaug~\citep{xu2024physaug}.
All experiments are conducted on two NVIDIA A100 GPUs.

\begin{figure*}[!ht]
    \centering
    \includegraphics[width=\linewidth]{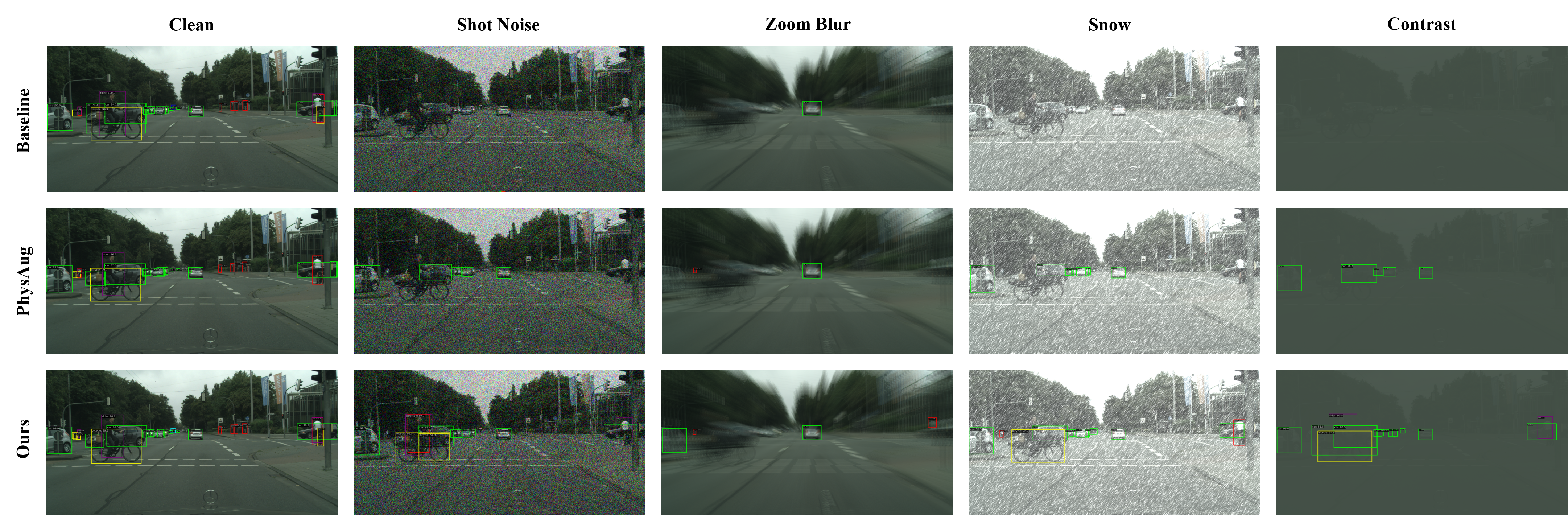}
    \caption{The visualizations of the baseline, PhysAug, and our model on the Cityscapes-C dataset are presented. These include results for the clean validation set, as well as specific corruptions such as shot noise, zoom blur, snow, and contrast. All images from the Cityscapes-C dataset are evaluated at a corruption level of 5.}
    \label{Figure 3}
\end{figure*}

\begin{figure}[!ht]
    \centering
    \begin{subfigure}[t]{0.48\linewidth}
        \centering
        \includegraphics[width=\linewidth]{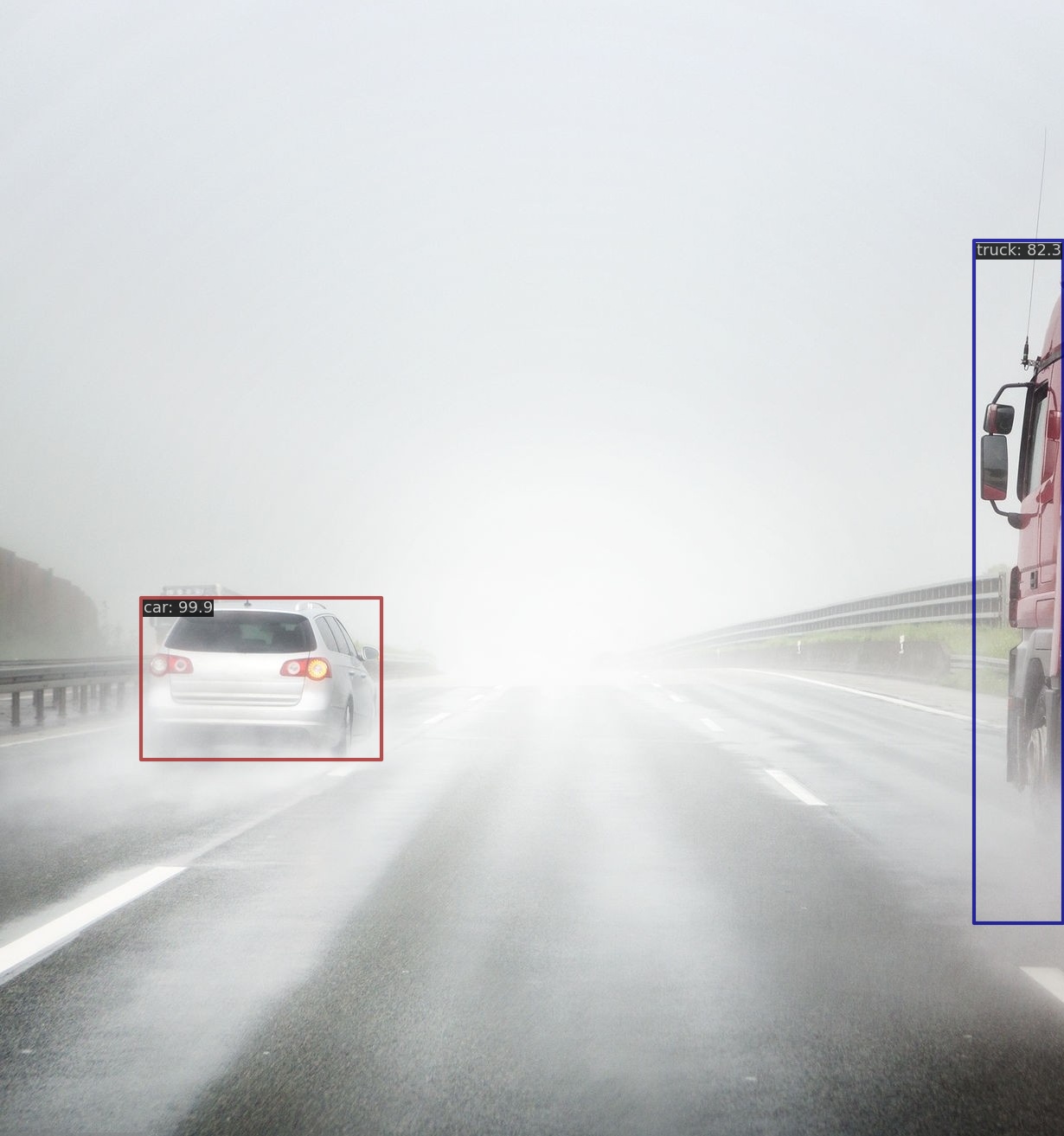}
        \caption{PhysAug}
        \label{fig:vis_main}
    \end{subfigure}
    \hfill
    \begin{subfigure}[t]{0.48\linewidth}
        \centering
        \includegraphics[width=\linewidth]{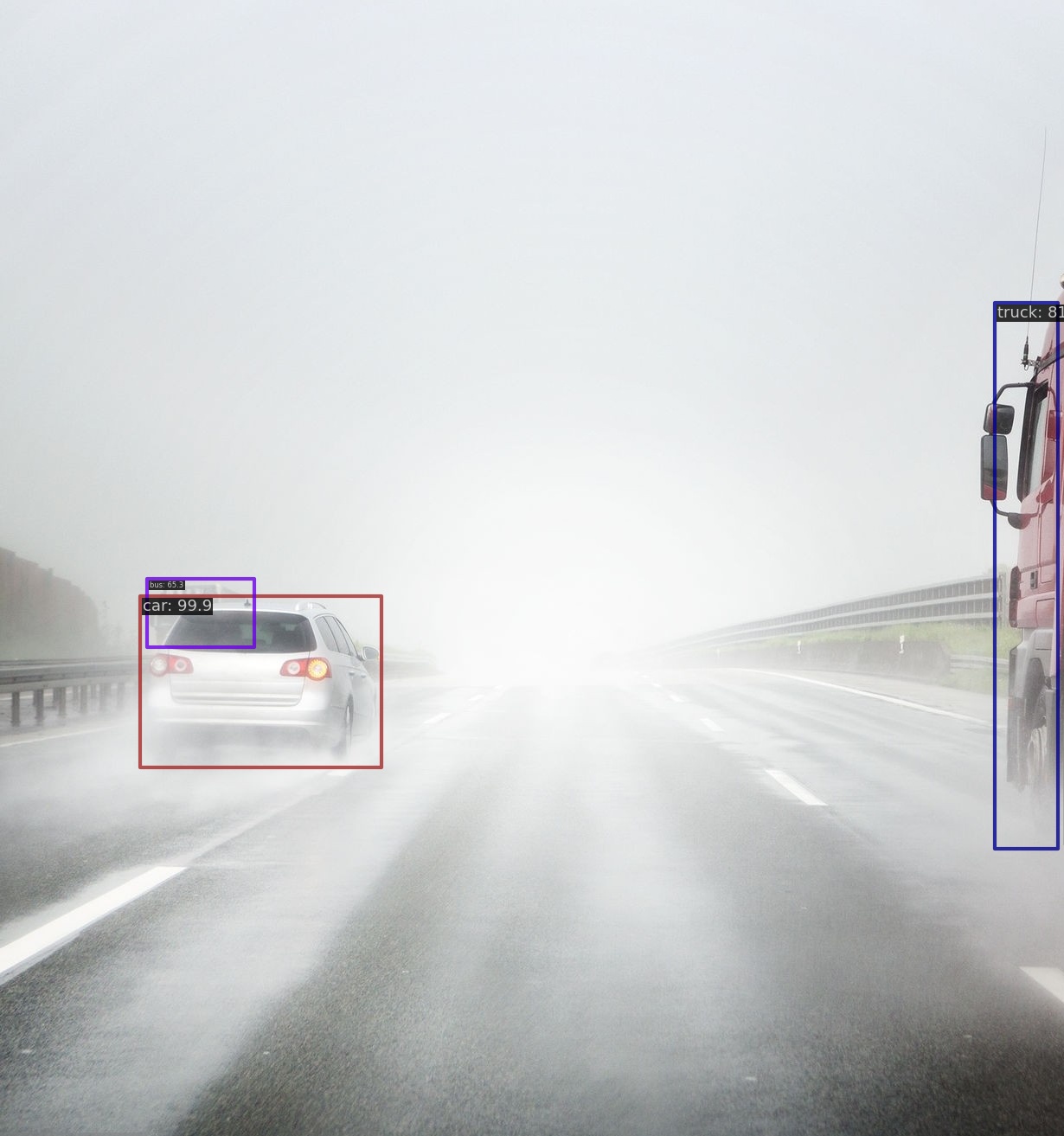} % 替换为你第二个子图的文件名
        \caption{Ours}
        \label{fig:vis_extra}
    \end{subfigure}
    \caption{Visualization of Daytime Foggy in DWD.}
    \label{dwd}
\end{figure}

\subsection{Performance on Robust Detection Benchmarks}
\subsubsection{Robustness Against Common Corruptions}  
To assess the effectiveness of our approach on Cityscapes-C, we compare our method with several Vision-only models and also investigate the potential of VLM-based models within this benchmark. We consider widely used data augmentation techniques, including Cutout~\citep{devries2017improved}, PhotoDistort~\citep{redmon2018yolov3}, AutoAug~\citep{zoph2020learning}, AugMix~\citep{hendrycks2019augmix}, StylizedAug~\citep{geirhos2018imagenet}, and PhysAug~\citep{xu2024physaug}. Additionally, we include methods that incorporate augmentation with loss function optimization, such as SupCon~\citep{khosla2020supervised}, FSCE~\citep{sun2021fsce}, and OA-DG~\citep{lee2024object}.  
While previous studies have primarily focused on Vision-only approaches, we take the first step in exploring the effectiveness of vision-language models (VLMs) in this benchmark. These models leverage textual representations to improve robustness under corruption. 
The results, summarized in Table~\ref{Table 1}, demonstrate that our approach consistently ranks among the top performers, achieving either the best or second-best results across various settings, highlighting its generalization ability. Notably, our method establishes a new state-of-the-art, outperforming the baseline detector by an average of $8.8$~mPC. Furthermore, it surpasses the previous best-performing method by an additional $1.6$~mPC, further validating its effectiveness.  

% \begin{figure*}[!ht]
%     \centering
%     \includegraphics[width=\linewidth]{visualization.png}
%     \caption{The visualizations of the baseline, PhysAug, and our model on the Cityscapes-C dataset are presented. These include results for the clean validation set, as well as specific corruptions such as shot noise, zoom blur, snow, and contrast. All images from the Cityscapes-C dataset are evaluated at a corruption level of 5.}
%     \label{Figure 3}
% \end{figure*}

% \begin{figure}[!ht]
%     \centering
%     \begin{subfigure}[t]{0.48\linewidth}
%         \centering
%         \includegraphics[width=\linewidth]{mist-115p.jpg}
%         \caption{PhysAug}
%         \label{fig:vis_main}
%     \end{subfigure}
%     \hfill
%     \begin{subfigure}[t]{0.48\linewidth}
%         \centering
%         \includegraphics[width=\linewidth]{mist-115x.jpg} % 替换为你第二个子图的文件名
%         \caption{Ours}
%         \label{fig:vis_extra}
%     \end{subfigure}
%     \caption{Visualization of Daytime Foggy in DWD.}
%     \label{dwd}
% \end{figure}

\begin{figure*}[!ht]
    \centering
    % 第一个子图
    \begin{subfigure}[b]{0.48\linewidth} % 子图宽度占总宽度的 48%
        \includegraphics[width=\linewidth]{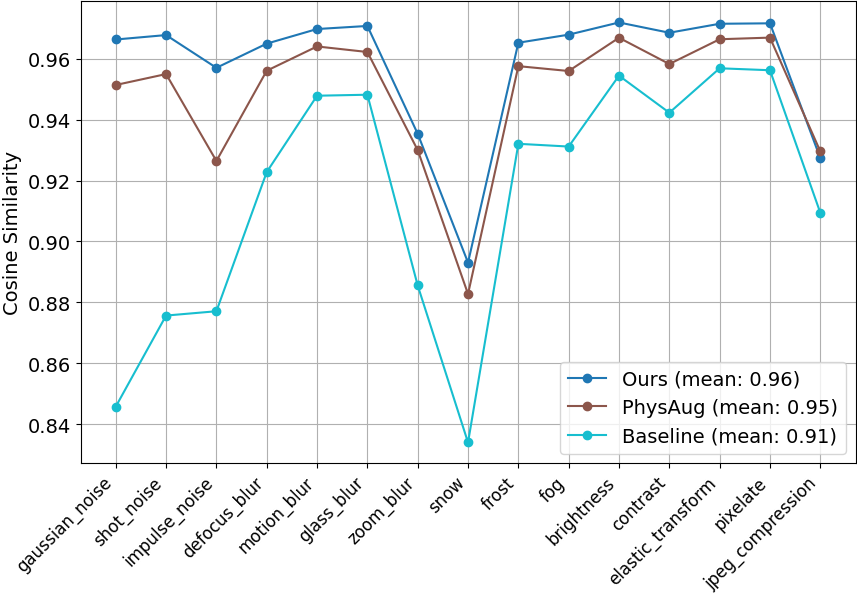}
        \caption{Severity 1.}
        \label{fig:cosine_similarity_1}
    \end{subfigure}
    \hfill % 两个子图之间的水平间距
    % 第二个子图
    \begin{subfigure}[b]{0.48\linewidth} % 子图宽度占总宽度的 48%
        \includegraphics[width=\linewidth]{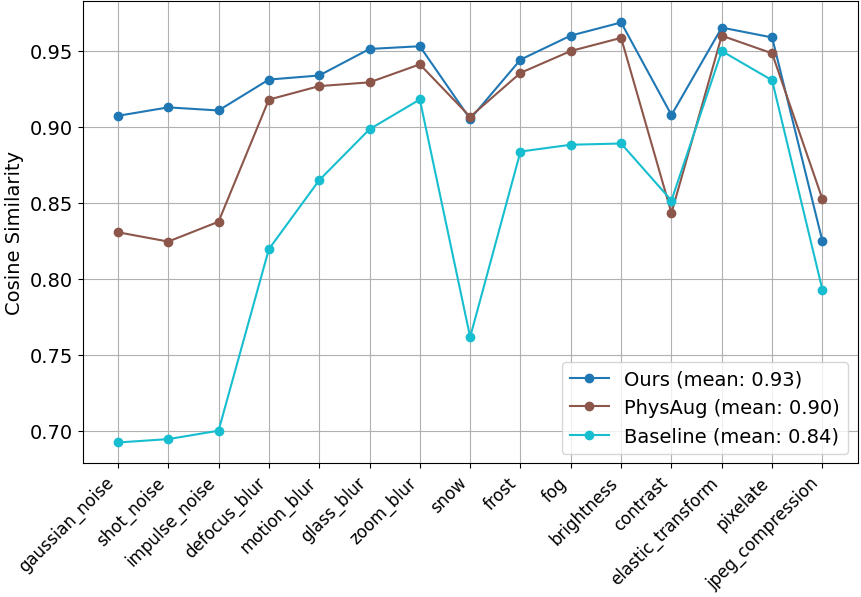} % 替换为你的第二张图文件名
        \caption{Severity 5.}
        \label{fig:cosine_similarity_2}
    \end{subfigure}
    % 整体标题
    \caption{The plots show cosine similarity between the Cityscapes validation set and Cityscapes-C at severity 1 and severity 5, with higher scores indicating better feature alignment between source and target features.}
    \label{Figure 4}
\end{figure*}

\subsubsection{Robustness Against Diverse Weather}
To further assess the effectiveness of our method in real-world scenarios, we use the DWD dataset as an additional benchmark. We compare our approach with state-of-the-art methods, categorized into Vision-only and VLM-based methods. Vision-only methods include SW~\citep{pan2019switchable}, IBN-Net~\citep{pan2019switchable}, IterNorm~\citep{huang2019iterative}, ISW~\citep{choi2021robustnet}, SHADE~\citep{zhao2022style}, OA-DG~\citep{lee2024object}, DivAlign~\citep{danish2024improving}, UFR~\citep{liu2024unbiased}, and PhysAug~\citep{xu2024physaug}. For VLM-based methods, we include CLIP-Gap~\citep{vidit2023clip}, Li's method~\citep{li2024prompt} and VLTDet~\citep{hummer2024strong}. The results of these comparisons, reported in Table~\ref{Table 2}, showcase their performance across five different scenarios. As shown in the table, our method performs the best in the night-sunny and dusk-rainy environments, surpassing all other methods. Moreover, compared to other methods that also use VLM, our approach better leverages the role of VLM in enhancing the model's generalization ability. Overall, our method outperforms the baseline by 7.9\%~mPC and surpasses the previous state-of-the-art by 0.6\%~mPC on this benchmark.

\begin{table}
  \caption{Ablation study on Cityscapes-C evaluating the performance of CRFI and CPRM. Corruption types per group ($N_C$): Noise (3), Blur (4), Weather (4), Digital (4), Total (15).}
  \label{tab:3}
  \fontsize{10}{12}\selectfont
  \centering
  \begin{tabular}{c|c|c|c} % 4列
    \hline
    Corruption & CRFI & CPRM & mPC  \\
    \hline
    \multirow{3}{*}{Noise}  
              & \ding{55}      & \ding{55}      & 0.9  \\
              & \ding{51}      & \ding{55}      & 17.4(\textcolor{red}{+16.5}) \\
              & \ding{51}      & \ding{51}      & 17.8(\textcolor{green}{+16.9}) \\
    \hline
    \multirow{3}{*}{Blur}  
              & \ding{55}      & \ding{55}      & 13.5 \\
              & \ding{51}      & \ding{55}      & 18.7(\textcolor{red}{+5.2}) \\
              & \ding{51}      & \ding{51}      & 19.0(\textcolor{green}{+5.5}) \\
    \hline
    \multirow{3}{*}{Weather}  
              & \ding{55}      & \ding{55}      & 19.1 \\
              & \ding{51}      & \ding{55}      & 27.3(\textcolor{red}{+8.2}) \\
              & \ding{51}      & \ding{51}      & 28.2(\textcolor{green}{+9.1}) \\
    \hline
    \multirow{3}{*}{Digital}  
              & \ding{55}      & \ding{55}      & 24.5 \\
              & \ding{51}      & \ding{55}      & 29.7(\textcolor{red}{+5.2}) \\
              & \ding{51}      & \ding{51}      & 30.3(\textcolor{green}{+5.8}) \\
    \hline
    \multirow{3}{*}{Total}  
              & \ding{55}      & \ding{55}      & 15.4\\
              & \ding{51}      & \ding{55}      & 23.9(\textcolor{red}{+8.5}) \\
              & \ding{51}      & \ding{51}      & 24.2(\textcolor{green}{+8.8}) \\
    \hline
  \end{tabular}
\end{table}

\subsection{Ablation Study}
We present an ablation study on the robustness of the model’s detection performance for Cross-modal and Region-aware Feature Interaction and Cross-domain Proposal Refining and Mixing on Cityscapes-C, as shown in Tab.~\ref{tab:3}.

% When Cross-Modal and Region-Aware Feature Interaction is incorporated into the model, a slight decline in performance is observed on the clean validation set, while a significant improvement is achieved on the corrupted validation set, with the greatest enhancement observed on the noise validation set.
% This indicates that Cross-Modal and Region-Aware Feature Interaction enhances robustness even at the cost of a lower overall accuracy. The performance degradation on the clean validation set also diminishes the robustness gains achieved by our method.

The results show that incorporating CRFI improves mPC scores significantly compared to both unmodified versions of the models. This suggests that integrating region-aware feature interactions between multimodal data helps enhance the model's ability to handle corruptions.
Furthermore, when CPRM is added to the modified version of the model using CRFI, there is another noticeable increase in the mPC score. This indicates that refining proposals from various domains can improve the overall robustness against corruption even after applying CRFI.

Overall, these findings demonstrate that combining regional-level interaction within modalities along with proposal refinement across domains contributes positively towards improving the model's resilience against image corruptions. It also highlights the importance of learning domain invariant features through the connection among diverse regions' vision-language knowledge, thus showcasing the effectiveness and rationale of our approach.

\section{Analysis}
To better understand the effectiveness of our method, we provide visualizations that highlight how our model extracts domain-invariant features. Specifically, we present:
(1) qualitative results on images with model-predicted bounding boxes.
(2) feature space visualizations to demonstrate the improved domain-invariant feature extraction capability.
In the following subsection, we present visualization results demonstrating our method's effectiveness in adverse weather and its impact on detection robustness.

\subsection{Visualization Analysis}
In Fig.~\ref{Figure 3}, we provide visualized object detection examples of the baseline method, PhysAug, and our proposed method under adverse environment conditions. 
In the clean validation set, the detection performance of each model is similar. However, when tested in noisy environments, the detection performance of all models decreases. The baseline model experiences a sharp decline in performance, with no detections in the shot noise, snow, and contrast conditions. PhysAug, as the previous state-of-the-art method, maintains relatively robust detection performance under various noisy conditions; however, it still suffers from occasional detection failures. Compared to the baseline and PhysAug, our approach demonstrates a marked improvement in detecting objects accurately. 
Furthermore, we visualize the detection results of our model under real-world adverse weather conditions by evaluating the Daytime Foggy subset of the DWD test set, as shown in Fig.~\ref{dwd}. Compared with PhysAug, our method can detect small or easily overlooked objects under adverse weather conditions.

\subsection{Feature Analysis}
To further demonstrate the effectiveness of our approach in learning domain-invariant representations, we analyze the feature similarities across different domains using cosine similarity, a common metric for measuring feature similarity. We extract features from both the source and target domains and compute their pairwise cosine similarity to assess the model's ability to extract domain-invariant features.
% We present the results as a line plot, as shown in Fig.~\ref{Figure 4}. Compared to the baseline and PhysAug, our approach consistently achieves higher cosine similarity across diverse datasets. 
We visualize the results in Fig.~\ref{Figure 4} as a line plot. Compared with the baseline and PhysAug, our method consistently achieves higher cosine similarity across diverse datasets under both the mildest and most severe corruption levels, indicating its superior ability to extract domain-invariant features.

\section{Conclusion}
% In this paper, we propose a method that consists of Cross-Modal and Region-Aware Feature Interaction and Cross-Domain Proposal Refining and Mixing. Cross-Modal and Region-Aware Feature Interaction to facilitate cross-modal domain-invariant feature learning. To further improve the localization performance under domain shifts, we introduce Cross-Domain Proposal Refining and Mixing,  which not only enhances the localization of regions of interest but also refines proposals across domains, enabling more robust region proposal learning. By integrating these two highly coupled techniques, our method effectively captures cross-modal domain-invariant features, enabling the detector to maintain high performance even in complex environments.
% \textcolor{red}{In this work, we introduced a novel vision-language invariant feature learning approach tailored for Single-Domain Generalized Object Detection, significantly advancing the ability to generalize across diverse unseen domains. Central to our method is the Cross-Modal and Region-Aware Feature Interaction mechanism, which dynamically integrates fine-grained visual and textual features to achieve robust intra- and inter-modal invariance at the region level. Additionally, our Cross-Domain Proposal Refining and Mixing strategy enhances localization by aligning and diversifying region proposals across domains. Experimental results on the Cityscapes-C and DWD benchmarks demonstrate that our approach achieves state-of-the-art performance, highlighting its potential for real-world multimedia applications.}

In this work, we introduced a novel cross-modal invariant feature learning approach tailored for Single-Domain Generalized Object Detection, significantly advancing the ability to generalize across diverse unseen domains. Central to our method is the Cross-modal and Region-aware Feature Interaction mechanism, which dynamically integrates fine-grained visual and textual features to achieve robust intra- and inter-modal invariance at the region level. Additionally, our Cross-domain Proposal Refining and Mixing strategy enhances localization by aligning and diversifying region proposals across domains. Experimental results on the Cityscapes-C and DWD benchmarks demonstrate that our approach achieves state-of-the-art performance, highlighting its potential for real-world multimedia applications.

\bibliographystyle{ACM-Reference-Format}
\bibliography{sample-base}

\end{document}